# Speaker Identification using MFCC-Domain Support Vector Machine

S. M. Kamruzzaman[1], A. N. M. Rezaul Karim[2], Md. Saiful Islam[3] and Md. Emdadul Haque[1]

[1]Department of Information and Communication Engineering
[3]Department of Computer Science and Engineering
University of Rajshahi, Rajshahi-6205, Bangladesh
[2]Department of Computer Science and Engineering
International Islamic University Chittagong, Bangladesh
Email: smzaman@gmail.com

**Abstract:** Speech recognition and speaker identification are important for authentication and verification in security purpose, but they are difficult to achieve. Speaker identification methods can be divided into text-independent and text-dependent. This paper presents a technique of text-dependent speaker identification using MFCC-domain support vector machine (SVM). In this work, melfrequency cepstrum coefficients (MFCCs) and their statistical distribution properties are used as features, which will be inputs to the neural network. This work firstly used sequential minimum optimization (SMO) learning technique for SVM that improve performance over traditional techniques Chunking, Osuna. The cepstrum coefficients representing the speaker characteristics of a speech segment are computed by nonlinear filter bank analysis and discrete cosine transform. The speaker identification ability and convergence speed of the SVMs are investigated for different combinations of features. Extensive experimental results on several samples show the effectiveness of the proposed approach.

## 1. Introduction

Speaker identification has been the subject of active research for many years, and has many potential applications where propriety of information is a concern. Speaker identification is the process of automatically recognizing a speaker by machine using the speaker's voice [11]. The most common application of speaker identification systems is in access control, for example, access to a room or privileged information over the telephone [14]. Also it has a very useful usage for speaker adaptation in automatic speech recognition system.

Speaker recognition can be classified into identification and verification [13]. Speaker identification is the process of determining which registered speaker provides a given utterance. Speaker verification, on the other hand, is the process of accepting or rejecting the identity claim of a speaker. Speaker recognition methods can also be divided into text-independent and text-dependent methods [9]. In a text-independent system, speaker models capture characteristics of somebody's speech, which show up irrespective of what one is saying. In a text-dependent system, on the other hand, the recognition of the speaker's identity is based on his or her speaking one or more specific phrases, like passwords, card numbers, PIN codes, etc [2]. The choice of which technology to use is application-specific

The objective of this work is to design an efficient system for human speech recognition that is able to identify and verify human speech more accurately. This work presents a technique of text-dependent speaker identification using MFCC-domain support vector machine (SVM). Mel-frequency cepstrum coefficients (MFCCs) and their statistical distribution properties are used as features, which will be inputs to the neural network [8].

## 2. Voice processing

The purpose of this module is to convert the speech waveform to some type of parametric representation (at a considerably lower information rate). The speech signal is a slowly time varying signal (it is called quasi-stationary). When examined over a sufficiently short period of time (between 5 and 100 ms), its characteristics are fairly stationary. However, over long periods of time (on the order of 0.2s or more) the signal characteristics change to reflect the different speech sounds being spoken. Therefore, short-time spectral analysis is the most common way to characterize the speech signal. A wide range of possibilities exist for parametrically representing the speech signal for the speaker recognition task, such as linear prediction coding (LPC), mel-frequency cepstrum coefficients

---

**Corresponding Author:** S. M. Kamruzzaman, Department of Information and Communication Engineering, University of Rajshahi, Rajshahi-6205, Bangladesh, Email: smzaman@gmail.com

(MFCC), and others. MFCC is perhaps the best known and most popular, and this feature has been used in this paper. MFCCs are based on the known variation of the human ear's critical bandwidths with frequency. The MFCC technique makes use of two types of filter, namely, linearly spaced filters and logarithmically spaced filters. To capture the phonetically important characteristics of speech, signal is expressed in the Mel frequency scale. This scale has a linear frequency spacing below 1000 Hz and a logarithmic spacing above 1000 Hz. Normal speech waveform may vary from time to time depending on the physical condition of speakers' vocal cord. Rather than the speech waveforms themselves, MFFCs are less susceptible to the said variations [1] [4].

### 2.1. The MFCC processor

A block diagram of the structure of an MFCC processor is given in Figure 1. The speech input is recorded at a sampling rate of 22050Hz. This sampling frequency is chosen to minimize the effects of *aliasing* in the analog-to-digital conversion process.

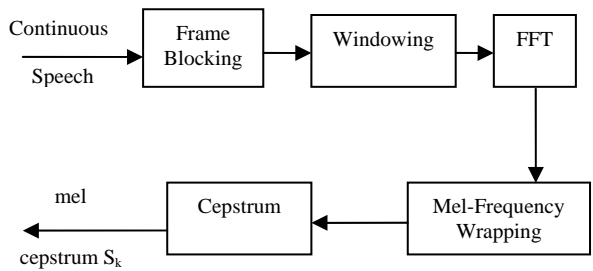

**Figure 1:** Block diagram of the MFCC processor.

### 2.2. Mel-frequency wrapping

The speech signal consists of tones with different frequencies. For each tone with an actual frequency, f, measured in Hz, a subjective pitch is measured on the 'Mel' scale. The *mel-frequency* scale is a linear frequency spacing below 1000Hz and a logarithmic spacing above 1000Hz. As a reference point, the pitch of a 1KHz tone, 40dB above the perceptual hearing threshold, is defined as 1000 mels. Therefore we can use the following formula to compute the mels for a given frequency *f* in Hz:

mel(f)= 2595*log10(1+f/700) ……….. (1)

One approach to simulating the subjective spectrum is to use a filter bank, one filter for each desired melfrequency component. The filter bank has a triangular bandpass frequency response, and the spacing as well as the bandwidth is determined by a constant mel-frequency interval.

### 2.3. Cepstrum

In the final step, the log mel spectrum has to be converted back to time. The result is called the mel frequency cepstrum coefficients (MFCCs). The cepstral representation of the speech spectrum provides a good representation of the local spectral properties of the signal for the given frame analysis. Because the mel spectrum coefficients are real numbers (and so are their logarithms), they may be converted to the time domain using the discrete cosine transform (DCT). The MFCCs may be calculated using the following equation [3] [5]:

$$\tilde{c}_n = \sum_{k=1}^{K} (\log \tilde{S}_k) \left[ n\left(k - \frac{1}{2}\right) \frac{\pi}{K} \right] \ldots\ldots\ldots(2)$$

where n=1, 2, …, K. The number of mel cepstrum coefficients, K, is typically chosen as 20. The first component, $\tilde{c}_0$, is excluded from the DCT since it represents the mean value of the input signal which carries little speaker specific information. By applying the procedure described above, for each speech frame of about 30 ms with overlap, a set of mel-frequency cepstrum coefficients is computed. This set of coefficients is called an *acoustic vector*. These acoustic vectors can be used to represent and recognize the voice characteristic of the speaker [7]. Therefore each input utterance is transformed into a sequence of acoustic vectors. The next section describes how these acoustic vectors can be used to represent and recognize the voice characteristic of a speaker.

## 3. Feature matching

The state-of-the-art feature matching techniques used in speaker recognition include, NETtalk, Time-Delay Neural Network (TDNN), Dynamic Time Warping (DTW), Hidden Markov Modeling (HMM), and Vector Quantization (VQ). A new approach, Support Vector Machine is used here, due to high accuracy.

### 3.1. Support vector machine

The support vector machine (SVM) is a linear machine pioneered by Vapnik [12]. Like multiplayer perceptron and radial basis function networks SVM can be used for pattern classification and nonlinear regression. The SVM become popular due to its attractive features and promising empirical performance. The idea of support vector machine is to construct a hyperplane as the decision surface in such a way that that the margin of separation between positive and negative example maximized. The SVM is an approximate implementation of the method of structural risk minimization (SRM). SRM principles have been shown to be superior to empirical risk minimization (ERM) principle employed by conventional neural networks. SRM minimize an upper bound on the expected risk as opposed to ERM that minimizes the error on the training data [6].

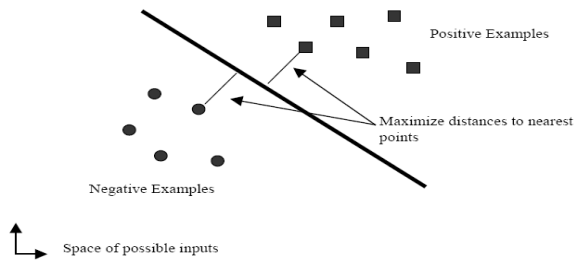

**Figure 2:** A linear support vector machine.

### 3.2. SVM training process

Though a solution for the quadratic optimization is guaranteed based on the KKT conditions, the number of computation required can be very high depending on the separability of the data and the number of training data points. Due to the size, the quadratic optimization problem that arises from SVMs cannot be easily solved via standard QP techniques. The quadratic form involves a matrix that has a number of elements equal to the square of the number of training examples. This matrix cannot be fit into 128 Megabytes if there are more than 4000 training examples. Vapnik [15] describes a method to solve the SVM quadratic problem, which has since been known as "chunking." The chunking algorithm uses the fact that the value of the quadratic form is the same if you remove the rows and columns of the matrix that corresponds to zero Lagrange multipliers. Therefore, the large quadratic optimization problem can be broken down into a series of smaller quadratic problems, whose ultimate goal is to identify all of the non-zero Lagrange multipliers and discard all of the zero Lagrange multipliers. At every step, chunking solves a quadratic optimization problem that consists of the following examples: every non-zero Lagrange multiplier from the last step, and the M worst examples that violate the KKT conditions [4], for some value of M. If there are fewer than M examples that violate the KKT conditions at a step, all of the violating examples are added in. Each quadratic sub-problem is initialized with the results of the previous sub-problem. At the last step, the entire set of non-zero Lagrange multipliers has been identified, hence the last step solves the large problem. Chunking seriously reduces the size of the matrix from the number of training examples squared to approximately the number of non-zero Lagrange multipliers squared. However, chunking still cannot handle large-scale training problems, since even this reduced matrix cannot fit into memory.

In 1997, Osuna, et al. [10] proved a theorem which suggests a whole new set of quadratic algorithms for SVMs. The theorem proves that the large quadratic optimization problem can be broken down into a series of smaller quadratic sub-problems. As long as at least one example that violates the KKT conditions is added to the examples for the previous sub-problem, each step will reduce the overall objective function and maintain a feasible point that obeys all of the constraints. Therefore, as sequence of quadratic sub-problems that always add at least one violator will be guaranteed to converge. Notice that the chunking algorithm obeys the conditions of the theorem, and hence will converge.

Osuna, et al. suggests keeping a constant size matrix for every quadratic sub-problem, which implies adding and deleting the same number of examples at every step [10]. Using a constant-size matrix will allow the training on arbitrarily sized data sets. The algorithm given in Osuna's paper [10] suggests adding one example and subtracting one example every step. Clearly this would be inefficient, because it would use an entire numerical quadratic optimization step to cause one training example to obey the KKT conditions. In practice, researchers add and subtract multiple examples according to unpublished heuristics [10]. In any event, a numerical QP solver is required for all of these methods. Numerical quadratic optimization is notoriously tricky to get right; there are many numerical precision issues that need to be addressed.

For each method, three steps are illustrated. The horizontal thin line at every step represents the training set, while the thick boxes represent the Lagrange multipliers being optimized at that step. For chunking, a fixed number of examples are added every step, while the zero Lagrange multipliers are discarded at every step.

Thus, the number of examples trained per step tends to grow. For Osuna's algorithm, a fixed number of examples are optimized every step: the same number of examples is added to and discarded from the problem at every step. For SMO, only two examples are analytically optimized at every step, so that each step is very fast.

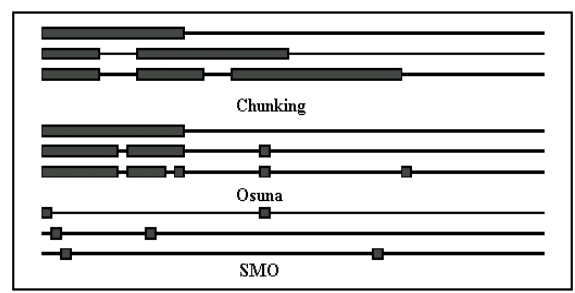

**Figure 3:** Three alternative methods for training SVMs: Chunking, Osuna's algorithm, and SMO.

### 3.3. Sequential minimal optimization (SMO)

Sequential minimal optimization (SMO) is a simple algorithm that can quickly solve the SVM problem without any extra matrix storage and without using numerical quadratic optimization steps at all. SMO decomposes the overall quadratic problem into quadratic sub-problems, using Osuna's theorem to ensure convergence.

Unlike the previous methods, SMO chooses to solve the smallest possible optimization problem at every step. For the standard SVM quadratic problem, the smallest possible optimization problem involves two Lagrange multipliers, because the Lagrange multipliers must obey a linear Equality constraint. At every step, SMO chooses two Lagrange multipliers to jointly optimize, finds the optimal values for these multipliers, and updates the SVM to reflect the new optimal values. The advantage of SMO lies in the fact that solving for two Lagrange multipliers can be done analytically. Thus, numerical quadratic optimization is avoided entirely. The inner loop of the algorithm can be expressed in a short amount of C code, rather than invoking an entire quadratic library routine. Even though more optimization sub-problems are solved in the course of the algorithm, each sub-problem is so fast that the overall quadratic problem is solved quickly.

In addition, SMO requires no extra matrix storage at all. Thus, very large SVM training problem scan fit inside of the memory of an ordinary personal computer or workstation. Because no matrix algorithms are used in SMO, it is less susceptible to numerical precision problems. There are two components to SMO an analytic method for solving for the two Lagrange multipliers, and a heuristic for choosing which multipliers to optimize.

### 4. Performance measure of SMO

The data set used to test SMO's speed was the UCI "adult" data set, which is available at [16]. There are 9888 examples in the "adult" training set. Two different SVMs were trained on the problem: a linear SVM, and a non-linear SVM. The timings for SMO versus chunking and Osuna for a linear SVM and non-linear SVM are shown in the table below:

**Table 1**: Timing of SMO versus chunking for a linear SVM and non-linear SVM.

| Training Set Size | For Linear SVM | | For Non-Linear SVM | |
|---|---|---|---|---|
| | SMO | Chunking | SMO | Chunking |
| 2477 | 2.2 | 13.1 | 26.3 | 64.9 |
| 3470 | 4.9 | 16.1 | 44.1 | 110.4 |
| 4912 | 8.1 | 40.6 | 83.6 | 372.5 |
| 7366 | 12.7 | 140.7 | 156.7 | 545.5 |
| 9888 | 24.7 | 239.3 | 248.1 | 907.6 |

For the linear SVM on this data set, the SMO training time faster than Chunking. This experiment is another situation where SMO is superior to Osuna and chunking in computation time. In this case, the scaling for SMO is somewhat better than chunking: SMO is a factor of between two and six times faster than chunking. The non-linear test shows that SMO is still faster than chunking. For the real-world test sets, SMO can be a factor of 1200 times faster for linear SVMs and a factor of 15 times faster for non-linear SVMs. Because of its ease of use and better scaling with training set size, SMO is a strong candidate for becoming the standard SVM training algorithm.

### 5. Experiment results

To show the effectiveness of the proposed approach some experimental results are shown. The experiment has done by testing the system in different ways and then comparing their result to estimate the final result. Here we used the sample of word "Zero" 20 times for each speaker. We work on total 8 speakers. All the data samples are collected from an international database. Our experiment results are given below by using Chunking and SMO training algorithm in SVM. Now, we will see how the system acts when using Chunking training algorithm in SVM.

**Table 2:** Results obtain from using Chunking training algorithm.

| Speaker ID | Total Sample | No. of correct samples | No. of false samples | Success rate in percentage (%) |
|---|---|---|---|---|
| 1 | 20 | 16 | 4 | 75 |
| 2 | 20 | 20 | 0 | 90 |
| 3 | 20 | 20 | 0 | 90 |
| 4 | 20 | 17 | 3 | 80 |
| 5 | 20 | 20 | 0 | 90 |
| 6 | 20 | 16 | 4 | 75 |
| 7 | 20 | 18 | 2 | 90 |
| 8 | 20 | 20 | 0 | 90 |

Now, we will see how the system acts when using sequential minimum optimization (SMO) training algorithm in SVM.

**Table 3:** Results obtain from using SMO training algorithm.

| Speaker ID | Total Sample | No. of correct samples | No. of false samples | Success rate in percentage (%) |
|---|---|---|---|---|
| 1 | 20 | 18 | 2 | 90 |
| 2 | 20 | 20 | 0 | 100 |
| 3 | 20 | 20 | 0 | 100 |
| 4 | 20 | 19 | 1 | 95 |
| 5 | 20 | 20 | 0 | 100 |
| 6 | 20 | 18 | 2 | 90 |
| 7 | 20 | 18 | 2 | 90 |
| 8 | 20 | 19 | 1 | 95 |

Here is the final result of our proposed approach

Table 4: Success rate in both SVM training methods.

| Training algorithm | No. of correct samples | No. of false samples | Success rate in percentage (%) |
|---|---|---|---|
| Chunking | 147 | 13 | 91.88 |
| SMO | 152 | 8 | 95 |

## 6. Conclusions

This paper successfully presents an approach based on neural networks for voice recognition and user identification. This approach based on MFCC-domain support vector machine by using SMO learning method. We have found a good result after testing the voice dependent system where there is a 91.88% success rate for using Chunking SVM training method and 95% for using SMO SVM training method. Thus we can conclude that, our approach have identified and verified human voice better than previous approaches.

The MFCC technique has been applied for speaker identification. VQ is used to minimize the data of the extracted feature. The study reveals that as number of centroids increases, identification rate of the system increases. It has been found that combination of Mel frequency and Hamming window gives the best performance. It also suggests that in order to obtain satisfactory result, the number of centroids has to be increased as the number of speakers increases. The study shows that the linear scale can also have a reasonable identification rate if a comparatively higher number of centroid is used. However, the recognition rate using a linear scale would be much lower if the number of speakers increases. Mel scale is also less vulnerable to the changes of speaker's vocal cord in course of time. The present study is still ongoing, which may include following further works. HMM may be used to improve the efficiency and precision of the segmentation to deal with crosstalk, laughter and uncharacteristic speech sounds. A more effective normalization algorithm can be adopted on extracted parametric representations of the acoustic signal, which would improve the identification rate further. Finally, a combination of features (MFCC, LPC, LPCC, Formant etc) may be used to implement a robust parametric representation for speaker identification.

## References

[1] Lawrence Rabiner and Biing-Hwang Juang, Fundamental of Speech Recognition, Prentice-Hall, Englewood Cliffs, N.J., 1993.

[2] Zhong-Xuan, Yuan & Bo-Ling, Xu & Chong-Zhi, Yu., Binary Quantization of Feature Vectors for Robust Text-Independent Speaker Identification, in IEEE Transactions on Speech and Audio Processing, Vol. 7, No. 1, January 1999.

[3] F. Soong, E. Rosenberg, B. Juang, and L. Rabiner, A Vector Quantization Approach to Speaker Recognition, AT&T Technical Journal, vol. 66, pp. 14-26, March/April 1987.

[4] Comp.speech Frequently Asked Questions WWW site, http://svr-www.eng.cam.ac.uk/comp.speech.

[5] Jr., J. D., Hansen, J., and Proakis, J., Discrete-Time Processing of Speech Signals, second ed. IEEE Press, New York, 2000.

[6] Aravind Ganapathiraju, Support Vector Machines for Speech Recognition, Ph. D dissertation, Mississippi State University, Mississippi, January 2002.

[7] Burges, C. J. C., A Tutorial on Support Vector Machines for Pattern Recognition, submitted to Data Mining and Knowledge Discovery, http://svm.research.bell-labs.com/SVMdoc.html, 1998.

[8] F. Jelinek, Statistical Methods for Speech Recognition, MIT Press, Cambridge, Massachusetts, USA, 1997.

[9] L.R. Rabiner and B.H. Juang, Fundamentals of Speech Recognition, Prentice Hall, Englewood Cliffs, New Jersey, USA, 1993.

[10] Osuna, E., Freund, R., Girosi, F., Improved Training Algorithm for Support Vector Machines, Proc. IEEE NNSP '97, 1997.

[11] Rabiner, L.R., A tutorial on Hidden Markov Models, Proceeding of IEEE, Vol.73, pp.1349-1387, 1989.

[12] Simon Haykin, McMaster University, Hamilton, Ontario, Canada, Neural Networks a Comprehensive Foundation, 2nd edition, pp.256-347.

[13] Squires, B and Sammut, Automatic Speaker Recognition: An application for Machine learning, Twelfth International Conference on Machine Learning, 1999.

[14] U. Bodenhausen, S. Manke, and A. Waibel, Connectionist Architectural Learning for High Performance Character and Speech Recognition, Proceedings of the International Conference on Acoustics Speech and Signal Processing, vol. 1, pp. 625-628, Minneapolis, MN, USA, April 1993.

[15] V. Vapnik, Estimation of Dependences Based on Empirical Data, Springer-Verlag, 1982.

[16] ftp://ftp.ics.uci.edu/pub/machine-learning-databases/adult.